\pdfoutput=1
\documentclass[runningheads]{llncs}
\usepackage{adjustbox}
\usepackage{graphicx}
\usepackage{multirow}
\usepackage{amsmath,graphicx}
\usepackage{color}
\usepackage{upgreek}
\usepackage{bbm}
\usepackage[utf8]{inputenc}
\usepackage{url}

\definecolor{cb_orange}{rgb}{1.0,0.51,0.0}
\definecolor{cb_blue}{rgb}{0.22,0.49,0.72}
\definecolor{cb_green}{rgb}{0.3,0.67,0.29}
\definecolor{cb_red}{rgb}{0.89,0.1,0.11}
\definecolor{cb_pink}{rgb}{1, 0, 0.4}

\begin{document}
\title{Histopathology Image Classification using\\ Deep Manifold Contrastive Learning}
%
%
\titlerunning{Deep Manifold Contrastive Learning}
%
\author{Jing Wei Tan\inst{1} \and
Won-Ki Jeong\inst{1} }
%
%
\institute{Department of Computer Science and Engineering, Korea University, Seoul, South Korea \\
\email{\{jingwei\textunderscore{92},wkjeong\}@korea.ac.kr}}

\maketitle              
\begin{abstract}
%
%

%

Contrastive learning has gained popularity due to its robustness with good feature representation performance. 
However, cosine distance, the commonly used similarity metric in contrastive learning, is not well suited to represent the distance between two data points, especially on a nonlinear feature manifold.
Inspired by manifold learning, we propose a novel extension of contrastive learning that leverages geodesic distance between features as a similarity metric for histopathology whole slide image classification. 
To reduce the computational overhead in manifold learning, we propose geodesic-distance-based feature clustering for efficient contrastive loss evaluation using prototypes without time-consuming pairwise feature similarity comparison.
The efficacy of the proposed method is evaluated on two real-world histopathology image datasets. 
Results demonstrate that our method outperforms state-of-the-art cosine-distance-based contrastive learning methods.
%
\end{abstract}
\section{Introduction}

Whole slide image (WSI) classification is a crucial process to diagnose diseases in digital pathology. 
Owing to the huge size of a WSI, the conventional WSI classification process consists of patch decomposition and per-patch classification, followed by the aggregation of per-patch results using multiple instance learning (MIL) for the final per-slide decision~\cite{kather2019deep}.
MIL constructs \textit{bag-of-features (BoF)} that effectively handles imperfect patch labels, allowing weakly supervised learning using per-slide labels for WSI classification.
Although MIL does not require perfect per-patch label assignment, it is important to construct good feature vectors that are easily separated into different classes to make the classification more accurate. 
Therefore, extensive research has been conducted on metric and representation learning~\cite{dml_ppt,zheng2019hardness}  
aimed at developing improved feature representation.


\begin{figure}[t]
\centering
\includegraphics[width=1\textwidth]{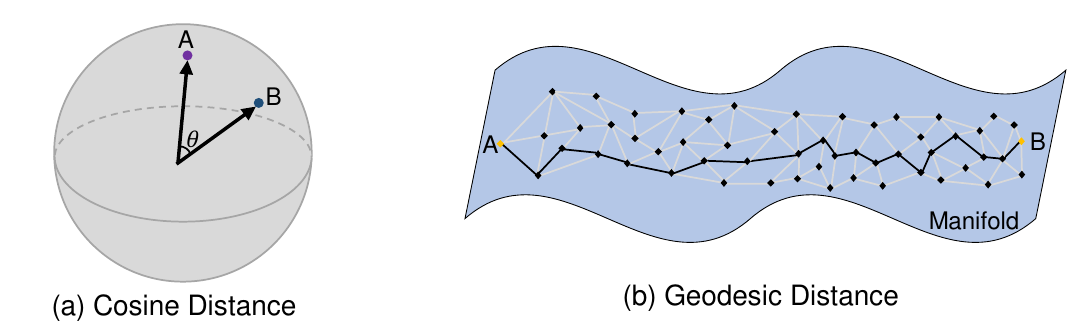}
\caption{Comparison of geodesic and cosine distance in n-dimensional space.}
\label{distance}
\end{figure} 

Recently, contrastive learning has demonstrated its robustness in the representational ability of the feature extractor, which employs self-supervised learning with a contrastive loss that forces samples from the same class to stay closer in the feature space (and vice versa). 
SimCLR~\cite{simclr} introduced the utilization of data augmentation and a learnable nonlinear transformation between the feature embedding and the contrastive loss to generally improve the quality of feature embedding. 
MoCo~\cite{moco} employed a dynamic dictionary along with a momentum encoder in the contrastive learning model 
to serve as an alternative to the supervised pre-trained ImageNet model in various computer vision tasks.
%
PCL~\cite{li2020prototypical} and HCSC~\cite{guo2022hcsc} integrated the k-means clustering and contrastive learning model by introducing prototypes as latent variables and assigning each sample to multiple prototypes to learn the hierarchical semantic structure of the dataset. 
These prior works used cosine distance as their distance measurement, which computes the angle between two feature vectors as shown in Fig.~\ref{distance}(a).
Although cosine distance is a commonly used distance metric in contrastive learning, we observed that the cosine distance approximates the difference between local neighbors and is insufficient to represent the distance between far-away points on a complicated, nonlinear manifold.

The main motivation of this work is to extend the current contrastive learning to represent the nonlinear feature manifold inspired by manifold learning. 
%
%
Owing to the manifold distribution hypothesis~\cite{lei2020geometric}, the relative distance between high-dimensional data is preserved on a low-dimensional manifold.
ISOMAP~\cite{theodoridis2009feature} is a well-known manifold learning approach that represents the manifold structure by using geodesic distance (i.e., the shortest path length between points on the manifold). 
There are several previous works that use manifold learning for image classification and reconstruction tasks, such as Lu~\textit{et al.}~\cite{lu2015multi} and Zhu~\textit{et al.}~\cite{zhu2018image}. 
However, the use of geodesic distance on the feature manifold for image classification is a recent development.
Aziere~\textit{et al.}~\cite{aziere2019ensemble} applied the random walk algorithm on the nearest neighbor graph to compute the pairwise geodesic distance and proposed the N-pair loss to maximize the similarity between samples from the same class for image retrieval and clustering applications. 
Gong~\textit{et al.}~\cite{gong2021deep} employed the geodesic distance computed using the Dijkstra algorithm on the k-nearest neighbor graph to measure the correlation between the original samples and then further divided each class into sub-classes to deal with the problems of high spectral dimension and channel redundancy in the hyperspectral images. 
However, this method captured the nonlinear data manifold structure on the original data (not on the feature vectors) only once at the beginning stage, which is not updated in the further training process.

In this study, we propose a hybrid method that combines manifold learning and contrastive learning to generate a good feature extractor (encoder) for histopathology image classification. 
Our method uses the sub-classes and prototypes as in conventional contrastive learning, but we propose the use of geodesic distance in generating the sub-classes to represent the non-linear feature manifold more accurately. 
By doing this, we achieve better separation between features with large margins, resulting in improved MIL classification performance.
%
The main contributions of our work can be summarized as follows: 
\begin{itemize}
\item We introduce a novel integration of manifold geodesic distance in contrastive learning, which results in better feature representation for the non-linear feature manifold.
We demonstrate that the proposed method outperforms conventional cosine-distance-based contrastive learning methods.
\item We propose a geodesic-distance-based feature clustering for efficient contrastive loss evaluation using prototypes without brute-force pairwise feature similarity comparison while approximating the overall manifold geometry well, which results in reduced computation.
%
%
\item We demonstrate that the proposed method outperforms other state-of-the-art (SOTA) methods with a much smaller number of sub-classes without complicated prototype assignment (e.g., hierarchical clustering).
\end{itemize}
To the best of our knowledge, this work is the first attempt to leverage manifold geodesic distance in contrastive learning for histopathology WSI classification.

\section{Method}
\begin{figure}[t]
\centering
\includegraphics[width=\columnwidth]{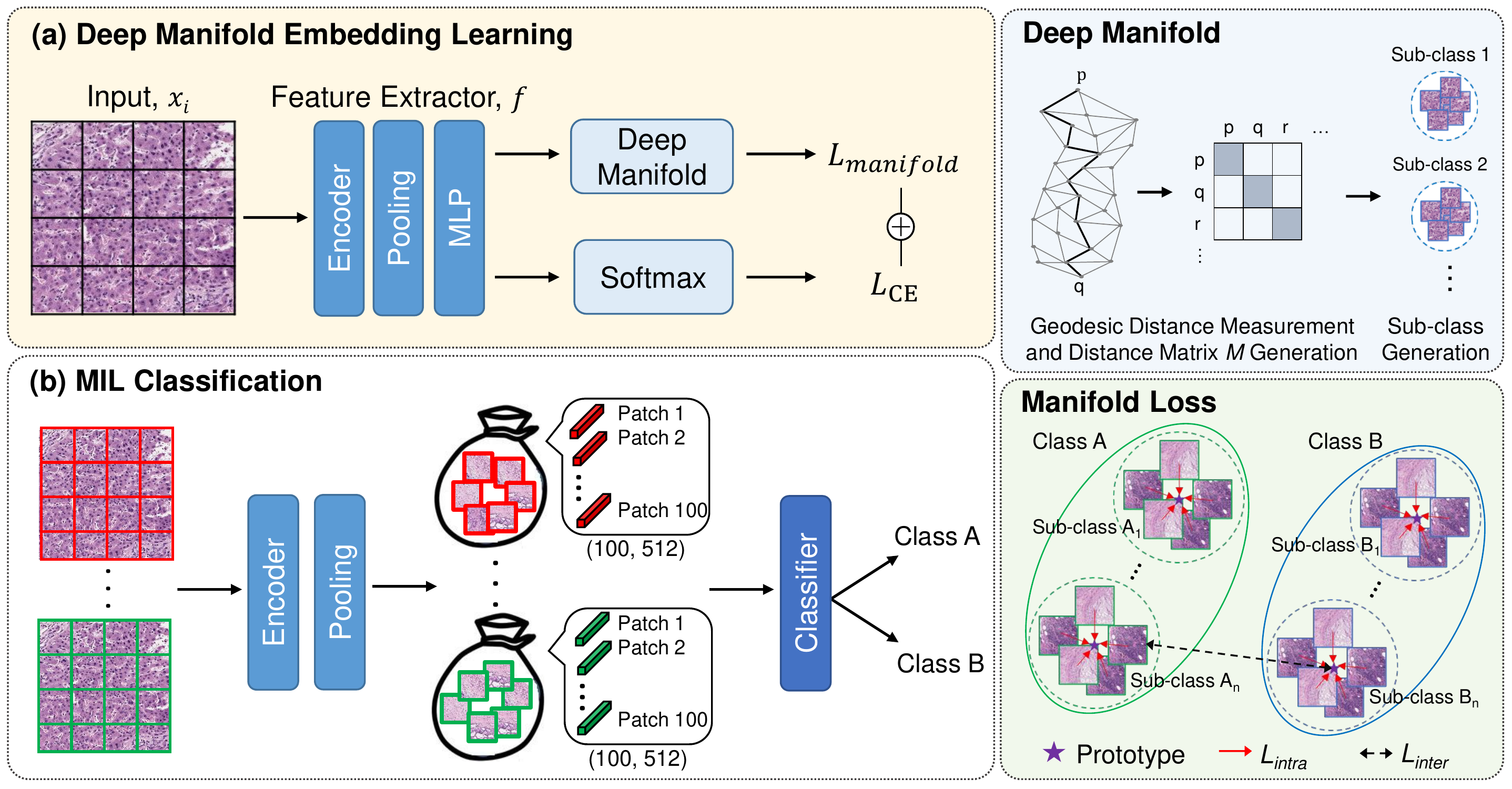}
\caption{Overview 
of our proposed method, which is composed of two stages: (a) deep manifold embedding learning and (b) MIL classification.}
\label{method}
\end{figure}

The overview 
of our proposed model is illustrated in Fig.~\ref{method}. 
It is composed of two stages: (1) train the feature extractor using deep manifold embedding learning and (2) train the WSI classifier using the deep manifold embedding extracted from the first stage. 
The input WSIs are pre-processed to extract $256\times256\times3$ dimensional patches from the tumor area at a 10x magnification level. 
Patches with less than $50\%$ tissue coverage are excluded from the experiment.

\subsection{Deep Manifold Embedding Learning}
As illustrated in Fig.~\ref{method}(a), we first feed the patches into a feature extractor \textit{$f$}, which is composed of an encoder, a pooling layer, and a multi-perceptron layer. 
The output is then passed through two different paths, namely, deep manifold and softmax paths.

\subsubsection{Deep manifold.} 
In this stage, the patches from each class are further grouped into sub-classes based on manifold geodesic distance. 
%
%
First, an undirected nearest neighbor graph $\textit{$G_c = (V_c, E_c)$}$ is constructed, where \textit{$V_c$} is a set of nodes made from the patch feature of the \textit{$c$}-th class extracted by $f$, and \textit{$E_c$} is the set of edges in the graph. 
Each node (patch feature) is connecting to its \textit{$k$}-nearest neighbors (KNN) based on the weighted edges computed with Euclidean distance, given that the neighbor samples on the manifold should have a higher potential to be in the same sub-class. 
The geodesic distance matrix \textit{$M$} on the manifold is then computed between each sample pair by using Dijkstra's algorithm based on the \textit{$G_c$}. 
The samples of each class are then further clustered into several sub-classes 
with agglomerative clustering based on the geodesic distance. 
In agglomerative clustering, 
all the patch features are initially treated as individual clusters and the nearest two individual clusters are merged based on the geodesic distance matrix to form new clusters. 
The distance matrix \textit{$M$} is then updated with the newly formed clusters. 
With the updated distance matrix \textit{$M$}, the nearest two individual clusters are again merged to form new clusters. 
These steps are 
repeated until the desired number $\textit{n}$ of sub-classes is achieved. 
%

\subsubsection{Loss functions.}
For the deep manifold training, we adopted two losses: (1) intra-subclass loss $\textit{$L_{intra}$}$ and (2) inter-subclass loss $\textit{$L_{inter}$}$. The main idea in intra-subclass loss is to make the samples from the same sub-class stay near their respective sub-class prototype. $\textit{$L_{intra}$}$ is formulated as follows:
\begin{equation}\label{eq1}
L_{intra} = \frac{1}{J \cdot I}\sum\limits_{j=1}^J\sum\limits_{i=1}^I(\textit{$f(x_j^i)-p^+$})^T(\textit{$f(x_j^i)-p^+$}) 
\end{equation}
where \textit{$x_j^i$} is the $i$-th patch in the $j$-th batch, \textit{$J$} represents the total number of batches, \textit{$I$} represents the total number of patches per batch, \textit{$f(\cdot)$} is the feature extractor, and \textit{$p^+$} indicates the positive prototype of the patch (i.e., the prototype of the subclass containing $x_j^i$). 
The prototype of each sub-class is computed by simply taking the mean of all the patch features that belong to each sub-class. 
Inter-subclass loss $\textit{$L_{inter}$}$ is proposed to make the sub-classes from a different class far apart from one another. The formulation of $\textit{$L_{inter}$}$ is as shown below:
\begin{equation}\label{eq2}
L_{inter} =\frac{1}{J}\sum\limits_{j=1}^J(\textit{$\triangle - {D}({f(Q_j^A)},{P^B}$})) 
\end{equation}
\begin{equation}\label{eq3}
D(Y,Z) = 
\max
\{
\sup_{y\in Y} d(y,Z), \sup_{z\in Z} d(z,Y)
\} 
\end{equation}
where \textit{$f(Q_j^A)$} is a set 
of patch features in batch $j$ from class \textit{$A$}, \textit{$P^B$} is a set of prototypes from the sub-classes of class \textit{$B$}, and \textit{$\triangle$} is a positive margin between classes on data manifold. 
\textit{$D(\cdot)$} is the Hausdorff distance, where \textit{sup} indicates supremum, \textit{inf} indicates infimum, and 
\( \mathrm{\textit{d}}{\textit{$(t,R)$}}=\underset{r\in R}\inf\vert\vert{\textit{t}-\textit{r}}\vert\vert \) 
which measures the distance from a data point \textit{$t\in Y$} to the subset \textit{$R \subseteq Y$}. 
Then, the manifold loss is formulated as
\begin{equation}\label{eq4}
L_{manifold} = L_{intra} + L_{inter} 
\end{equation}

Another path via softmax is simply trained on outputs from the feature extractor with the ground truth slide-level labels \textit{y} by the cross-entropy loss \textit{$L_{CE}$}, which is defined as follows: 
%
\begin{equation}\label{eq5}
L_{CE} = -\frac{1}{J \cdot I}\sum\limits_{j=1}^J \sum\limits_{i=1}^I \textit{$y_j^i$} \cdot\log\textit{$\hat{y}_j^i$} + (1-y_j^i) \cdot \log (1-\textit{$\hat{y}_j^i$})  
\end{equation}
where \textit{y} is the ground truth slide-level label and \textit{$\hat{y}$} is predicted label.

Finally, the total loss for the first stage is defined as follows: 
\begin{equation}\label{eq6}
L_{total} = L_{manifold} + L_{CE} 
\end{equation}

\subsection{MIL Classification}
As illustrated in Fig.~\ref{method}(b), in the second stage, the pre-trained feature extractor from the previous stage is then deployed to extract features for bag generation. 
A total of 50 bags are generated for each WSI, in which each bag is composed of the concatenation of the features from 100 patches in 512 dimensions. 
These bags are fed into a classifier with two layers of multiple perceptron layers (512 neurons) and a Softmax layer and then trained with a binary cross-entropy loss. 
After the classification, majority voting is applied to the predicted labels of the bags to derive the final predicted label for each WSI.

\section{Result}

\subsection{Datasets}
We tested our proposed method on two different tasks: (1) intrahepatic cholangiocarcinomas(IHCCs) subtype classification and (2) liver cancer type classification. 
%
The dataset for the former task was collected from 168 patients with 332 WSIs from (anonymized) hospital. 
IHCCs can be further categorized into small duct type (SDT) and large duct type (LDT).
%
Using gene mutation information as prior knowledge, we collected WSIs with wild KRAS and mutated IDH genes for use as training samples in SDT, and WSIs with mutated KRAS and wild IDH genes for use in LDT. 
The rest of the WSIs were used as testing samples.
%
The liver cancer dataset for the latter task was composed of 323 WSIs, in which the WSIs can be further classified into hepatocellular carcinomas (HCCs) (collected from Pathology AI Platform~\cite{paip}) and IHCCs.
We collected 121 WSIs for the training set, and the remaining WSIs were used as the testing set.

\subsection{Implementation Detail}

We used a pre-trained VGG16 with ImageNet as the initial encoder, which was further modified via deep manifold model training using the proposed manifold and cross-entropy loss functions. 
The number of nearest neighbors \textit{$k$} and the number of sub-classes \textit{$n$} were set to 5 and 10, respectively. 
In the deep manifold embedding learning model, the learning rates were set to 1e-4 with a decay rate of 1e-6 for the IHCCs subtype classification and to 1e-5 with a decay rate of 1e-8 for the liver cancer type classification. 
The $k$-nearest neighbors graph and the geodesic distance matrix are updated once every five training epochs, which is empirically chosen to balance running time and accuracy. 
%
%
To train the MIL classifier, we set the learning rate to 1e-3 and the decay rate to 1e-6.
We used batch sizes 64 and 4 for training the deep manifold embedding learning model and the MIL classification model, respectively. 
The number of epochs for the deep manifold embedding learning model was 50, while 50 and 200 epochs for the IHCCs subtype classification and liver cancer type classification, respectively. 
As for the optimizer, we used stochastic gradient decay for both stages.
The result shown in the tables is the average result from 10 iterations of the MIL classification model.


\subsection{Experimental Results}

The performance of different models from two different datasets is reported in this section. 
For the baseline model, we chose the pre-trained VGG16 feature extractor with an MIL classifier, which is the same as our proposed model except that the encoder is retrained using the proposed loss. 
Two SOTA methods using contrastive learning and clustering, PCL~\cite{li2020prototypical} and HCSC~\cite{guo2022hcsc}, are compared with our method in this study. 
The MIL classification result of the IHCCs subtype classification is shown in Table~\ref{tab1}. 
Our proposed method outperformed the baseline CNN by about 4\% increment in accuracy, precision, recall, and F1 score. 
Note that our method only used 20 sub-classes but outperformed PCL (using 2300 sub-classes) by 4\% and HCSC (using 112 sub-classes) by 5\% in accuracy.


\begin{table} [tb]
\centering
\caption{Classification performance on IHCCs subtype and liver cancer type dataset. (Acc.: Accuracy, Prec.: Precision, Rec.: Recall, F1: F1 Score, NA: Not Applicable)}\label{tab1}
\begin{tabular}{l|c|cccc|cccc} \hline
\multirow{2}{*}{Method} & \multirow{1}{*}{Prototype} & \multicolumn{4}{c|}{IHCC Subtype} & \multicolumn{4}{c}{Liver Cancer Type}\\ \cline{3-10}
\multicolumn{1}{c|}{} &\multicolumn{1}{c|}{Number}& Acc. & Prec. & Rec. & F1 & Acc. & Prec. & Rec. & F1  \\ \hline \hline
CNN & NA &0.7315 & 0.7372 & 0.7315 & 0.7270 & 0.7710&0.7781&	0.7719	&0.7657\\ 
PCL & 500-800-1000 &0.7386 & 0.7478 & 0.7394 & 0.7354&0.8146 &0.7898 & 0.8146 & 0.7979 \\ 
HCSC & 2-10-100 &0.7230 &  0.7265 & 0.7230 & 0.7231 & 0.7995 &\textbf{0.8524}&0.7995&0.7825 \\ \hline 
Ours & 20 &\textbf{0.7703} & \textbf{0.7710} & \textbf{0.7678} & \textbf{0.7668}& \textbf{0.8239} & 0.8351 & \textbf{0.8239} & \textbf{0.8227}  \\\hline \hline
\end{tabular}
\end{table}


The result of liver cancer type classification is also shown in Table~\ref{tab1}. 
Our method achieved about 5\% improvement in accuracy against the baseline and 1\% to 2\%  improvement in accuracy against the SOTA methods. 
%
Moreover, it outperformed the SOTA methods with 
far fewer prototypes and without complicated hierarchical prototype assignments.
To further evaluate the effect of prototypes, we conducted an ablation study for different prototype assignment strategies as shown in Table~\ref{tab5}. 
%
%
Here, global prototypes imply assigning a single prototype per class while local prototypes imply assigning multiple prototypes per class (one per sub-class). 
When both are used together, it 
implies a hierarchical prototype assignment where local prototypes 
interact with the corresponding global prototype. 
As shown in this result, the model with local prototypes only performed about 4\% higher than did the model with global prototypes only.  
Meanwhile, the combination of both prototypes achieved a similar performance to that of the model with local prototypes only.  
Since the hierarchical (global + local) assignment did not show a significant improvement but instead increased computation, we used only local sub-class prototypes in our final experiment setting.


\begin{table} [tb]
\centering
\vspace{-2mm}
\caption{Ablation study of prototype assignment strategies.}\label{tab5}
\begin{tabular}{l | c | c  c c c} \hline
Prototype & Prototype Number & Accuracy & Precision & Recall & F1 Score   \\ \hline \hline 
Global &  2& 0.7365 &	0.7390 &	0.7365&	0.7353\\ 
Local &  20 & \textbf{ 0.7703}&	0.7710&0.7678&0.7668\\ 
Global + Local & 22 (20 + 2)& 0.7698& \textbf{0.7735}&\textbf{0.7698}& \textbf{0.7692}\\\hline \hline
\end{tabular}
\vspace{-2mm}
\end{table}

\begin{table} [tb]
\centering
\caption{Classification performance of geodesic distance and cosine distance.}\label{tab3}
\begin{tabular}{l | c | c  c c c} \hline
Method & Number of sub-classes & Accuracy & Precision & Recall & F1 Score  \\ \hline \hline
Cosine distance & 2 & 0.7519 & 0.7552 & 0.7519 & 0.7503 \\
Cosine distance & 20 & 0.7576 & 0.7589 & 0.7576 & 0.7571 \\
Ours & 20 & \textbf{0.7703} & \textbf{0.7710} & \textbf{0.7678} & \textbf{0.7668}  \\\hline \hline
\end{tabular}
\vspace{-2mm}
\end{table} 

\begin{figure}[t]
\centering
\includegraphics[width=\columnwidth]{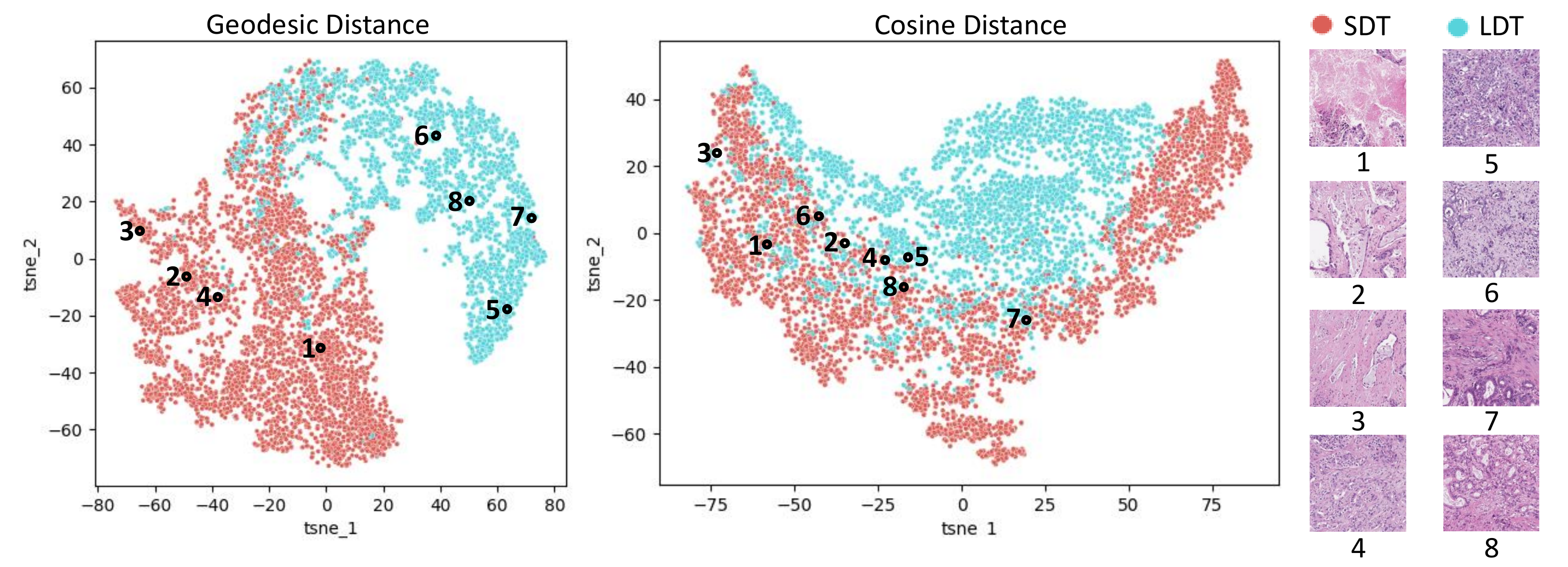}
\caption{Comparison of geodesic and cosine distance in feature space.(1)-(4) are the patches from SDT and (5)-(8) are the patches from LDT.}
\label{distance_comparison}
\vspace{-2mm}
\end{figure}

Since one of our contributions is the use of geodesic distance, we assessed the efficacy of the method by comparing it with the performance using cosine distance, as shown in Table~\ref{tab3}.
To measure the performance of the cosine-distance-based method, we simply replaced our proposed manifold loss with NT-Xent loss~\cite{chen2020simple}, which uses cosine distance in their feature similarity measurement. 
Two cosine distance experiments were conducted as follows: (1) use only their ground-truth class without further dividing the samples into sub-classes (i.e., global prototypes) and (2) divide the samples from each class into 10 sub-classes by using k-means clustering (i.e., local prototypes). 
As shown in Table~\ref{tab3}, using multiple local prototypes shows slightly better performance compared to using global prototypes.
By switching the NT-Xent loss with our geodesic-based manifold loss, the overall performance is increased by about 2\%. 
Fig.~\ref{distance_comparison} visually compares the effect of the geodesic and cosine distance-based losses. 
Two scatter plots are t-SNE projections of feature vectors from the encoders trained using geodesic distance and cosine distance, respectively.
Red dots represent SDT samples and blue dots represent LDT samples from the IHCCs dataset (corresponding histology thumbnail images are shown on the right). 
In this example, all eight cases are correctly classified by the method using geodesic distance while all cases are incorrectly classified by the method using cosine distance. 
It is clearly shown that geodesic distance can correctly measure the feature distance (similarity) on the manifold so that SDT and LDT groups are located far away in the t-SNE plot, whereas cosine distance failed to separate these groups and they are located nearby in the plot.

 
%
%
%
\subsection{Conclusion and Future Work}
In this paper, we proposed a novel geodesic-distance-based contrastive learning for histopathology image classification. 
Unlike conventional cosine-distance-based contrastive learning methods, our method can represent nonlinear feature manifold better and generate better discriminative features.
One limitation of the proposed method is the extra computation time for graph generation and pairwise distance computation using the Dijkstra algorithm. 
In the future, we plan to optimize the algorithm and apply our method to other datasets and tasks, such as multi-class classification problems and natural image datasets.

\newpage



\bibliographystyle{splncs04}
\nocite{*}
\bibliography{reference}

\end{document}